\documentclass[preprint,5p,times,twocolumn,authoryear]{./elsarticle}
\usepackage{amssymb}
\usepackage{amsmath}
\usepackage{multirow}
\usepackage{booktabs}
\usepackage{color}
\usepackage{lineno, hyperref}

\begin{document}

\begin{frontmatter}
\title{ST-FlowNet: An Efficient Spiking Neural Network for Event-Based Optical Flow Estimation}

\author[label1]{Hongze Sun}
\author[label1]{Jun Wang}
\author[label1]{Wuque Cai}
\author[label1,label2]{Duo Chen}
\author[label1]{Qianqian Liao}
\author[label1]{Jiayi He}
\author[label1,label3]{Yan Cui}
\author[label1,label4]{Dezhong Yao\corref{cor1}}
\author[label1,label4]{Daqing Guo\corref{cor1}}
\address[label1]{Clinical Hospital of Chengdu Brain Science Institute, MOE Key Lab for NeuroInformation, China-Cuba Belt and Road Joint Laboratory on Neurotechnology and Brain-Apparatus Communication, School of Life Science and Technology, University of Electronic Science and Technology of China, Chengdu 611731, China.}
\address[label2]{School of Artificial Intelligence, Chongqing University of Education, Chongqing 400065, China.}
\address[label3]{Sichuan Academy of Medical Sciences and Sichuan Provincial People's Hospital, Chengdu 610072, China.}
\address[label4]{Research Unit of NeuroInformation (2019RU035), Chinese Academy of Medical Sciences, Chengdu 611731, China.}
\cortext[cor1]{Corresponding authors: dyao@uestc.edu.cn (Dezhong Yao) and dqguo@uestc.edu.cn (Daqing Guo).}

\begin{abstract}
Spiking neural networks~(SNNs) have emerged as a promising tool for event-based optical flow estimation tasks due to their capability for spatio-temporal information processing and low-power computation. However, the performance of SNN models is often constrained, limiting their applications in real-world scenarios. To address this challenge, we propose ST-FlowNet, a novel neural network architecture specifically designed for optical flow estimation from event-based data. The ST-FlowNet architecture integrates ConvGRU modules to facilitate cross-modal feature augmentation and temporal alignment of the predicted optical flow, thereby improving the network's ability to capture complex motion patterns. Additionally, we introduce two strategies for deriving SNN models from pre-trained artificial neural networks~(ANNs): a standard ANN-to-SNN conversion pipeline and our proposed BISNN method. Notably, the BISNN method alleviates the complexities involved in selecting biologically inspired parameters, further enhancing the robustness of SNNs for optical flow estimation tasks. Extensive evaluations on three benchmark event-based datasets demonstrate that the SNN-based ST-FlowNet model outperforms state-of-the-art methods, achieving superior accuracy in optical flow estimation across a diverse range of dynamic visual scenes. Furthermore, the energy efficiency of models also underscores the potential of SNNs for practical deployment in energy-constrained environments. Overall, our work presents a novel framework for optical flow estimation using SNNs and event-based data, contributing to the advancement of neuromorphic vision applications.
\end{abstract}

\begin{keyword}
Spiking neural networks \sep Optical flow estimation \sep Event-based images \sep Training methods.
\end{keyword}
\end{frontmatter}

%%%%%%%%%%%%%%%%%%%%%%%%%%%%%%%%%%%%%%%%% Introduction %%%%%%%%%%%%%%%%%%%%%%%%%%%%%%%%%%%%%%%%%
\begin{figure*}[!t]
\centering
\includegraphics[width=7.2in]{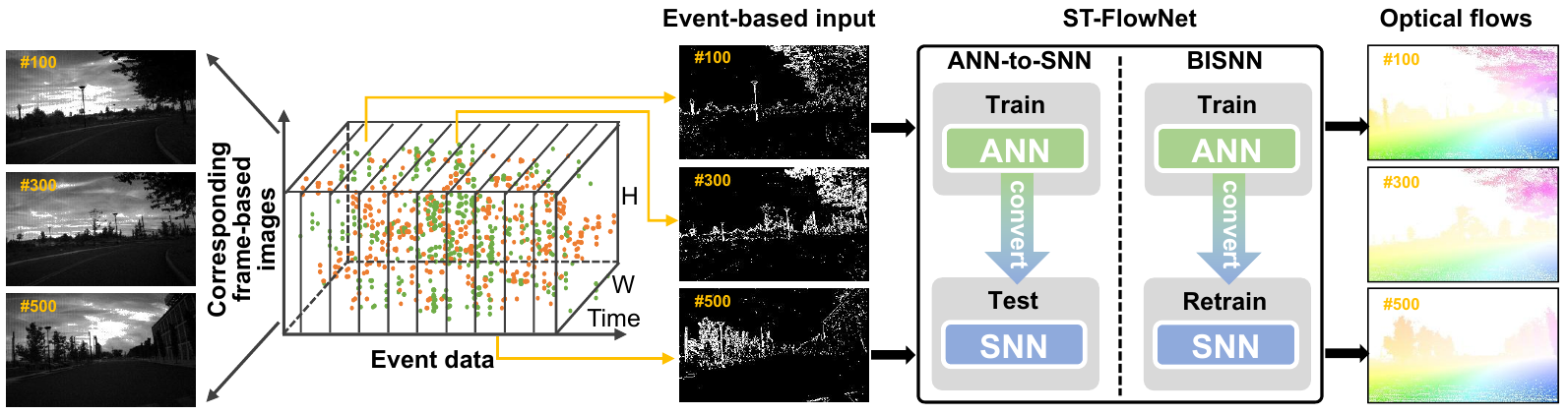}
\caption{The framework of our proposed optical flow estimation method is illustrated. The ST-FlowNet~(both ANN and SNN) models utilize event-based images as input data. Following training on the ANN model, an SNN ST-FlowNet model is derived through the A2S conversion or BISNN method. Optical flow prediction is achievable using both ANN and SNN models. Additionally, for reference, the corresponding frame-based images are presented in the left black box. Conventional frame-based images exhibit abundant spatial texture information indiscriminately, while event-based images emphasize motion-related objects by leveraging spatio-temporal cues simultaneously.}
\label{Figure1}
\end{figure*}

\section{Introduction}
As the third generation of neural networks~\citep{maass1997networks}, spiking neural networks (SNNs) have garnered increasing attention in recent years~\citep{zheng2021going}. Unlike conventional artificial neural networks (ANNs), SNNs employ bio-inspired neurons and discrete spike trains to mimic the complex spatiotemporal dynamics of the human brain~\citep{zheng2024temporal, sun2023synapse}. These characteristics enable SNNs to achieve competitive performance across a wide range of neuromorphic intelligent tasks~\citep{10047993, yu2023brain}. Furthermore, the event-driven information coding and transmission mechanisms ensure lower power consumption, thus enhancing the feasibility of SNNs for hardware implementation~\citep{yao2023sparser}.

Optical flow estimation is a fundamental topic in the field of computer vision and has extensive applications~\citep{ilg2017flownet, 9511282}, particularly in motion-related intelligent tasks. For instance, by clustering motion patterns in different regions, optical flow can assist object segmentation models in separating the foreground and background~\citep{zitnick2005consistent}. The motion vector of the target object is critically important for predicting the search space in object tracking tasks~\citep{10047993}. Optical flow also serves as a compensation tool for frame insertion-based image reconstruction and enhancement~\citep{fan2021optical, wang2020deep}. However, prevalent research has predominantly concentrated on optical flow estimation from frame-based images captured by conventional cameras, resulting in significant performance degradation in challenging scenarios characterized by high-speed motion or unfavorable lighting conditions~\citep{zhai2021optical}. Event-based neuromorphic cameras, capable of asynchronously recording changes in light intensity within a high dynamic range of illumination~\citep{han2020neuromorphic}, respond to these challenging scenarios effectively and exhibit potential for minimizing energy consumption, thereby presenting an attractive prospect for deployment on edge devices~\citep{yu2023brain}. Furthermore, drawing inspiration from inherent imaging mechanisms, event-based images can circumvent errors introduced by assumptions related to the conservation of pixel intensity in optical flow estimation~\citep{gaur2022lucas}. Thus, the advantages of event-based optical flow estimation are significant when compared with conventional approaches~\citep{zhu2018multivehicle, mueggler2017event, scheerlinck2020fast}.

Owing to the structural specificity inherent in the event modality, current approaches typically involve the reconstruction of event data into frame-based images for the estimation tasks handled by conventional ANN models~\citep{zhu2018ev, ilg2017flownet}. Despite some notable progress, these approaches often neglect potential errors inherent in the reconstruction process. Representatively, the resulting reconstructed images frequently exhibit pronounced motion blur in spatial visual cues, particularly as the sampling time window lengthens~\citep{stoffregen2020reducing, tian2022event}. Concurrently, the valuable dynamic features embedded in the temporal domain remain underutilized. In contrast, an SNN model, comprising spiking neurons as fundamental units, accepts spiking events as input and uses firing spikes as a medium for information propagation and presentation~\citep{cai2023spatial}. The SNN is an essential tool for event-based feature extraction~\citep{sun2023synapse}. Consequently, a natural solution to address the aforementioned challenges involves developing an SNN model specifically tailored for event-based optical flow estimation. 

Unlike conventional ANN models, training SNN models with standard backpropagation~(BP) is challenging due to the non-differentiable nature of spike signals~\citep{wu2018spatio}, leading to suboptimal performance in real-world applications. To address this challenge, several methods have been proposed, which can be classified into four main categories: (1) unsupervised learning inspired by biological neuronal plasticity~\citep{diehl2015unsupervised}, (2) indirect training via ANN-to-SNN conversion~(A2S)~\citep{deng2021optimal}, (3) spatio-temporal backpropagation~(STBP) employing surrogate gradient approximation~\citep{wu2018spatio}, and (4) hybrid training strategies~\citep{sun2023synapse}. Among these, STBP has emerged as the most widely adopted approach, enabling SNN training with a procedure similar to that of ANNs while maintaining competitive accuracy. However, STBP-based SNN models typically require extended temporal windows for effective training, resulting in significant computational overhead. Moreover, approximation errors introduced by surrogate gradients further constrain their performance in complex real-world tasks. Consequently, A2S methods have gained prominence as an alternative, allowing the derivation of SNN models from pre-trained ANNs. Compared to the STBP method, A2S strategy substantially reduces training complexity, particularly for intricate task-specific models. Additionally, hybrid strategies integrating diverse training paradigms or neuronal plasticity mechanisms have attracted growing interest, further enhancing SNN training efficiency and model performance.	

To fully exploit the potential of event data in optical flow estimation, it is imperative to effectively address two key challenges: (1) developing a novel architecture capable of simultaneous spatio-temporal feature extraction based on event data; and (2) proposing a novel model training strategy for achieving superior optical flow estimation performance. Thus far, limited research has delved into these pertinent issues. Therefore, this work introduces a novel method to effectively tackle these challenges~(shown in Fig.~\ref{Figure1}). We present our ST-FlowNet architecture tailored specifically to estimate event-based optical flow. By incorporating the ConvGRU layers~\citep{ballas2015delving}, ST-FlowNet achieves cross-scale fusion of dynamic optical flow features from event data in the temporal dimension. In contrast to prior models inspired by pyramidal architectures~\citep{ilg2017flownet}, ST-FlowNet employs a more streamlined decoder structure, enabling direct decoding of latent information across the entire multi-scale feature space. Following training on the ANN model, the SNN model for optical flow estimation can be derived through two strategies: the A2S method or our proposed bio-information-fused training strategy~(BISNN). Notably, our BISNN approach achieves parameter-free model conversion while preserving the performance of optical flow estimation.

Our key contributions in this paper are summarized as follows:  
\begin{itemize}
\item We propose ST-FlowNet, a novel architecture designed for efficient optical flow estimation by leveraging spatio-temporal features in event-based data.  
\item We present the first framework for constructing efficient SNN models for optical flow estimation, utilizing an A2S conversion approach.  
\item We introduce a novel parameter-free SNN training strategy, which further mitigates training challenges while enhancing overall training efficiency.  
\item Our experimental results demonstrate that ST-FlowNet attains superior performance when compared with other state-of-the-art models on challenging benchmark datasets such as MVSEC, ECD, and HQF.
\end{itemize}

%%%%%%%%%%%%%%%%%%%%%%%%%%%%%%%%%%%%%%%%% Related Work %%%%%%%%%%%%%%%%%%%%%%%%%%%%%%%%%%%%%%%%%
\section{Related Work}
Optical flow estimation is a fundamental task in the field of computer vision~\citep{ilg2017flownet}, garnering significant research attention. In this section, we initially delve into the evolution of event-based optical flow estimation. Subsequently, our focus shifts to the models built by SNNs, which demonstrate notable proficiency in extracting spatio-temporal features from event-based images.

\subsection{Event-based Optical Flow Estimation}
Considering the distinctive attributes of event images in contrast to the RGB modality, early research focused on innovating paradigms for event-based optical flow computation, but only achieved limited success in several simple scenarios~\citep{ benosman2013event}. The introduction of large-scale benchmark datasets has facilitated applying deep learning models to event-based optical flow estimation tasks~\citep{zhu2018multivehicle, mueggler2017event}, significantly enhancing the efficacy of optical flow estimation and reducing the challenges associated with model design~\citep{scheerlinck2020fast, tian2022event}. Previous research attempted to use encoder-decoder networks to decouple spatio-temporal features embedded in event data across multiple resolutions~\citep{zhu2018ev}. A series of variants emerged through the refinement of network structures and adjustments to loss functions, achieving reliable and accurate optical flow estimation facilitated by self-supervised and end-to-end learning methods~\citep{scheerlinck2020fast, tian2022event}. However, considering the high temporal resolution characteristics inherent in event data, models founded on general convolution structures struggle to comprehensively use the temporal features of the event data, thereby limiting the effectiveness of event-based optical flow estimation.

\subsection{SNN Models for Optical Flow Estimation}
Considering the inherent advantages of SNNs in spatio-temporal feature extraction~\citep{tavanaei2019deep}, our work focuses on models using spiking neurons which are naturally adept at capturing spatio-temporal visual cues embedded in event data for precise optical flow estimation. Recently, numerous models based on spiking neurons have been proposed for event-based optical flow estimation~\citep{paredes2019unsupervised,hagenaars2021self, zhang2023event}. Similar to traditional ANN models, shallow convolutional SNN models trained using spike-time-dependent-plasticity learning methods~\citep{diehl2015unsupervised} have been introduced, demonstrating promising performance for relatively simple tasks~\citep{paredes2019unsupervised}. To enhance SNN model performance in complex real-world scenarios, a logical approach involves constructing hybrid models that leverage the strengths of both ANNs and SNNs~\citep{diehl2015unsupervised}. Additionally, drawing inspiration from the BP-style direct training method for SNN models~\citep{wu2018spatio}, a series of fully spiking models have also attained state-of-the-art performance on event-based data when compared with advanced ANN models~\citep{hagenaars2021self, zhang2023event}. However, constrained by training approaches and network architectures, we posit that event-based optical flow estimation by SNN models can be further optimized by ANN conversion~\citep{deng2021optimal}.

\begin{figure*}[!t]
\centering
\includegraphics[width=7.2in]{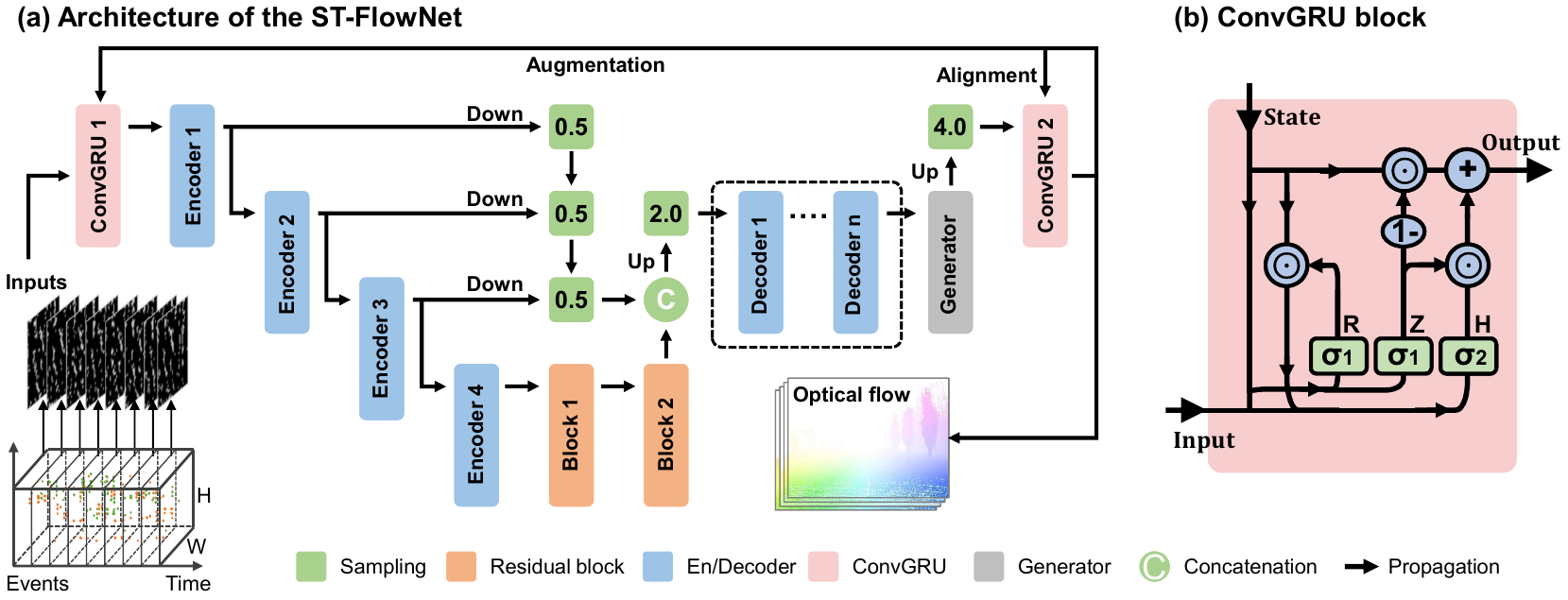}
\caption{(a) The ST-FlowNet architecture is illustrated. Following pre-processing by a ConvGRU layer, the enhanced event-based input undergoes downsampling via four encoder layers. The resulting minimal feature maps produced by encoder4 traverse two residual block layers, ensuring robust feature extraction. Through the concatenation of feature maps at various levels, decoder layers and a generator are deployed for basic optical flow prediction. Furthermore, the basic predicted optical flow is fed through a ConvGRU layer to fuse historical sequential temporal feature and generate the final predicted optical flow. (b) Schematic illustration depicting the architecture of ConvGRU. A ConvGRU unit integrates both current input and state information to produce a corresponding output. The symbol $\odot$ denotes the Hadamard product, and $\sigma$ is the activation function.}
\label{Figure2}
\end{figure*}

%%%%%%%%%%%%%%%%%%%%%%%%%%%%%%%%%%%%%%%%% Preliminaries %%%%%%%%%%%%%%%%%%%%%%%%%%%%%%%%%%%%%%%%%
\section{Preliminary}
To facilitate a clearer understanding of our work, this section provides a brief overview of SNNs.
To date, a variety of spiking neuron models have been developed to emulate the spatio-temporal dynamics of biological neurons. Among these, the leaky integrate-and-fire~(LIF) model has become the foundational computational unit in constructing SNNs, offering a balance between biological plausibility and computational efficiency~\citep{abbott1999lapicque}. The membrane potential dynamics of an LIF neuron are typically governed by the following differential equation:
\begin{equation}
\tau \frac{dv(t)}{dt} = V_{\text{rest}} - v(t) + I(t),
\label{eq_pre1}
\end{equation}
accompanied by the spike generation mechanism defined as:
\begin{equation}
s(t) =
\begin{cases}
1, & \text{if } v(t) \geq \theta, \\
0, & \text{otherwise},
\end{cases}
\label{eq_pre2}
\end{equation}
where \( v(t) \), \( I(t) \), and \( s(t) \) denote the membrane potential, input current, and output spike at time \( t \), respectively. The parameters \( \tau \), \( V_{\text{rest}} \), and \( \theta \) represent the membrane time constant, resting potential, and firing threshold. A spike is emitted when the membrane potential \( v(t) \) reaches or exceeds the threshold \( \theta \), after which the membrane potential is reset.

For implementation convenience, the continuous-time differential Eqs~\eqref{eq_pre1} and~\eqref{eq_pre2} are typically approximated in discrete-time form. The membrane potential update at each time step can be expressed as:
\begin{equation}
v(t) = \alpha(t) \cdot \tilde{v}(t-1) + \beta \cdot I(t),
\label{eq_pre3}
\end{equation}
where \( V_{\text{rest}} \) is assumed to be zero for simplicity. The parameters \( \alpha(t) \) and \( \beta \) control the decay of the membrane potential and the scaling of the input current, respectively. Here, \( \tilde{v}(t-1) \) represents the reset membrane potential at the previous time step, and is defined as:
\begin{equation}
\tilde{v}(t-1) = 
\begin{cases}
v(t-1) \cdot (1 - s(t-1)), & \text{for hard reset}, \\
v(t-1) - s(t-1) \cdot \theta, & \text{for soft reset},
\end{cases}
\label{eq_pre4}
\end{equation}
where \( s(t-1) \) indicates whether a spike was emitted at the previous time step. In the hard reset mechanism, the membrane potential is set to zero following a spike, while in the soft reset case, it is reduced by the threshold value \( \theta \).

%%%%%%%%%%%%%%%%%%%%%%%%%%%%%%%%%%%%%%%%% Method %%%%%%%%%%%%%%%%%%%%%%%%%%%%%%%%%%%%%%%%%
\section{Method}
We present an overview of the proposed ST-FlowNet model in Fig.~\ref{Figure2}. Specifically, ST-FlowNet builds on the foundation of the end-to-end FlowNet model~\citep{ilg2017flownet}, characterized by encoder and decoder layers. To effectively leverage the event modality, which inherently possesses rich spatio-temporal information, we incorporate ConvGRU layers~\citep{cho2014learning} for sequential feature argumentation and alignment. Additionally, we use SNN models derived from pre-trained ANN models to enhance the ST-FlowNet’s proficiency in spatio-temporal feature extraction. In this section, we elaborate on the event representation, network architecture, and training methodology used for ST-FlowNet.

\subsection{Event Representation}
In contrast to conventional frame-based images, which capture light intensity at discrete exposure times, event-based images~($\left \{ \left [ x_{k}, y_{k}, t_{k}, p_{k} \right ]  \right \}_{k=1}^{K} $) asynchronously record changes in light intensity at location $[x_{k}, y_{k}]$ along with their respective polarity~($p_{k}\in \left \{p^{+},  p^{-} \right \} $). Here, $K$ represents the total number of events, and $t_{k}$ denotes the timestamp of the $k$-th event. For the sake of convenience, the event inputs are aggregated into $N$ group of frame-based representations $f_{n}^{+}(\tilde{x} , \tilde{y})$ and $f_{n}^{-}(\tilde{x} , \tilde{y})$~($n = 1,...,N$), which are defined by the following formulas:
\begin{align}
\label{eq1} f_{n}^{+}(\tilde{x} , \tilde{y})&={\textstyle \sum_{i}^{}}p_{i}^{+}, \\
\label{eq2} f_{n}^{-}(\tilde{x} , \tilde{y})&={\textstyle \sum_{i}^{}}p_{i}^{-}, 
\end{align}
subject to $i = \frac{K}{N}(n-1)+1,...,\frac{K}{N}n$. The inputs to the models are presented as:
\begin{align}
\label{eq3} &\textrm{ANN:}[f_{1}^{+},f_{1}^{-},...,f_{N}^{+},f_{N}^{-}],\\
\label{eq4} &\textrm{SNN:}\{[f_{1}^{+},f_{1}^{-}],...,[f_{N}^{+},f_{N}^{-}]\}.
\end{align}
Concretely, we concatenate N groups of event-based images into a $2N$-channel data representation~(Eq.~\ref{eq3}), which serves as input to the ANN model, ensuring the complete information of the inputs is captured. For the SNN model, the value of N determines the temporal resolution of the raw data, and each pair of images is sequentially processed by the model~(Eq.~\ref{eq4}).

\subsection{Network Architecture}
As illustrated in Fig.~\ref{Figure2}, the ST-FlowNet model is structured in a semi-pyramidal encoder-decoder architecture~\citep{ilg2017flownet}. Diverging from prior approaches, we incorporate ConvGRU layers to enhance the model’s sequential feature representation. The ConvGRU layer retains its original architecture~(shown in Fig.~\ref{Figure2}(b)). However, the hidden state, which is used to modify information across different time steps, has been redesigned.

Specifically, to process the input data, a \textbf{ConvGRU} layer is employed to combined the present input ${I}_{t}$ with the inherent information $O_{t-1}$. The augmented input data $\tilde{I}_{t}$ is characterized as follows:
\begin{equation}
\tilde{I}_{t}=(1-Z_{t})\odot O_{t-1}+Z_{t}\odot H_{t}.
\label{eq5}
\end{equation}
$R_{t}$, $Z_{t}$ and $H_{t}$ represent the reset gate, update gate and candidate hidden state respectively, and are calculated as follows: 
\begin{align}
\label{eq6} R_{t}&=\mathrm{\sigma_{1} } (\mathrm{\xi_{R}} (O_{t-1},I_{t})),\\
\label{eq7} Z_{t}&=\mathrm{\sigma_{1} } (\mathrm{\xi_{Z}} (O_{t-1},I_{t})),\\
\label{eq8} H_{t}&=\mathrm{\sigma_{2}} (\mathrm{\xi_{I}} (R_{t}\odot  O_{t-1},I_{t})).
\end{align}
Here, $\mathrm{\xi_{R}}$, $\mathrm{\xi_{Z}}$ and $\mathrm{\xi_{I}}$ represent convolutional operations, $\mathrm{\sigma_{1}}$ and $\mathrm{\sigma_{2}}$ are the sigmoid and tanh activation functions. Intuitively, the candidate hidden state $H_{t}$ balances and retains pertinent information from the previous optical flow and current input~(Eq.\ref{eq8}). The reset gate $R_{t}$ governs the degree to which historical information is removed in $H_{t}$. Ultimately, the augmented input $\tilde{I}_{t}$ undergoes an update via a weighted summation of $O_{t-1}$ and $H_{t}$~(Eq.\ref{eq5}). $O_{-1}$ is established as $0$ at the initial time step $t=0$.

For every augmented input $\tilde{I_{t}}$, ST-FlowNet uses a convolutional architecture as \textbf{Encoders} to derive feature maps across multiple hierarchical levels, spanning from low-level to high-level representations~(denoted as $\{F_{t}^{l}\}_{l=1}^{4}$). The dimensions of the feature maps undergo a reduction by half in each subsequent layer, while the channel count experiences a twofold increment. Furthermore, the resultant highest-level representation $F_{t}^{4}$ undergoes processing via two residual \textbf{Block}, aimed at conducting a more in-depth exploration of the underlying deep features~(denoted as $\tilde{F_{t}^{4}}$). 

To maximize the exploitation of optical flow information across various spatial scales, our architecture concatenates all feature maps that have undergone downsampling to attain a uniform size. Subsequently, this concatenated representation is upsampled to yield an input for subsequent decoder stages. This distinctive procedure is formally written as follows:
\begin{equation}
F_{t} = \mathcal{U}_{2}(\mathrm{concate} (\mathcal{D}_{8}(F_{t}^{1}),\mathcal{D}_{4}(F_{t}^{2}),\mathcal{D}_{2}(F_{t}^{3}),\tilde{F_{t}^{4}})),
\label{eq9}
\end{equation}
where $\mathcal{D}$ and $\mathcal{U}$ represent the down and upsampling operations, respectively, with a subscript denoting the specific sampling factor. In our approach, we implement the downsampling process using convolutional layers with corresponding strides, while the upsampling process is achieved through bilinear interpolation.

After the feature extraction stage, tandem \textbf{Decoder} modules, each consisting of a convolutional layer with uniform input and output dimensions, decode the cross-scale feature representation into a predicted optical flow. The decoders are tasked with interpreting the flow information embedded in the feature space and generating a corresponding flow map. Finally, a \textbf{Generator} module, comprising a single convolutional layer with two filters, produces the basic predicted flow. In our method, the generator acts as a refinement stage, further enhancing the optical flow prediction generated by the decoders.

To further improve the optical flow prediction, we incorporate a \textbf{ConvGRU2} layer~\citep{cho2014learning} following the \textbf{Generator} module to output the final optical flow. \textbf{ConvGRU2} uses the predicted flow from the previous time step as its state and the basic predicted flow from the current time step as its input. This temporal integration of optical flow information enables ST-FlowNet to capture long-range dependencies in the flow sequence. In addition, the output $O_{t}$ of \textbf{ConvGRU2} is then used to augment the state information for \textbf{ConvGRU1} at the next time step, providing a more comprehensive representation of the optical flow dynamics.

\subsection{Model Training}
In this work, we use two types of cross-model transformation methodology, A2S conversion~\citep{bu2023optimal, deng2021optimal, rathi2020diet} and hybrid bio-information fusion training, to generate an optimal SNN for optical estimation while preserving the original ANN model's accuracy. 

\subsubsection{ANN-to-SNN conversion}
The fundamental process of the A2S conversion encompasses three key phases: first, constructing ANN and SNN models with identical architectures but employing distinct basic neurons; second, training ANN models using the standard BP method; and third, converting the trained ANN model into an SNN model, ensuring extensive retention of model accuracy.

In the ANN model, where information is represented with continuous values, the neuron is defined as follows:
\begin{equation}
a^{l}=h(w^{l} \cdot a^{l-1}),
\label{eq10}
\end{equation}
where $a^{l}$ is the output in the $l$-th layer, $w^{l}$ signifies the weight between the $(l-1)$-th and $l$-th layers, and $h(\cdot)$ denotes the activation function. In the SNN model, we use the LIF model~\citep{abbott1999lapicque} as the spiking neuron for network construction. Without loss of generality, the iterative form of the membrane potential $v^{l}(t)$ of the LIF neuron is described as follows~\citep{rathi2020diet}:
\begin{equation}
v^{l}(t)= e^{-\tau} \cdot v^{l}(t-1)-s^{l}(t-1)\cdot \theta ^{l} + w^{l}\cdot s^{l-1}(t)\cdot\theta ^{l-1}.
\label{eq11}
\end{equation}
Here, $\theta ^{l}$ is the spiking firing threshold, $\tau \in \left [ 0,+\infty  \right )$ represents the membrane potential decay factor, and $s^{l}$ denotes the spike output generated by:
\begin{equation}
s^{l}(t)=H(v^{l}(t)-\theta ^{l}).
\label{eq12}
\end{equation}
Using the Heaviside step function $H(\cdot)$, the LIF spiking neuron emits a spike once the membrane potential surpasses the predetermined firing threshold. To circumvent notable accuracy diminution, we use the quantization clip-floor activation function~\citep{bu2023optimal} in the ANN model instead of the ReLU function:
\begin{equation}
h(x)=\lambda ^{l}\cdot clip(\frac{1}{L}\left \lfloor \frac{x\cdot L}{\lambda ^{l}}   \right \rfloor,0,1  ).
\label{eq13}
\end{equation}
The clip function sets the upper bound to 1 and the lower bound 0. $\lfloor \cdot \rfloor$ denotes the floor function. Prior research has substantiated that the conversion process is theoretically lossless when the hyper-parameter $L$ aligns with the desired time windows $T$ of the SNN, and the trained parameter $\lambda^{l}$ corresponds to the spike firing threshold~\citep{bu2023optimal,deng2021optimal}.

The training process for the optical flow estimation network is conducted using the ANN model. Motivated by the limited availability of ground truth optical flow data, we train our optical flow estimation network through a self-supervised approach~\citep{hagenaars2021self}. The comprehensive loss function encompasses two fundamental components: a contrast loss $\mathcal{L}_{\mathrm{contrast}}$ and a smoothness loss $\mathcal{L}_{\mathrm{smooth}}$. The contrast loss $\mathcal{L}_{\mathrm{contrast}}$ uses a reformulated contrast maximization proxy loss to gauge the accuracy of optical flow estimation by assessing the motion compensation performance of an image reconstructed from the predicted optical flow~\citep{hagenaars2021self, mitrokhin2018event, zhu2019unsupervised}. The smoothness loss $\mathcal{L}_{\mathrm{smooth}}$ uses Charbonneir smoothness function~\citep{charbonnier1994two, zhu2018ev, zhu2019unsupervised} to regulate the optical flow variation between neighboring pixels. Consequently, the total loss is defined as:
\begin{equation}
\mathcal{L}_{\mathrm{flow}}=\mathcal{L}_{\mathrm{contrast}}+\lambda \mathcal{L}_{\mathrm{smooth}},
\label{eq14}
\end{equation}
where the scalar $\lambda$ balances the respective weights of the contrast loss and the smoothness loss. 

Once the ANN model is fully trained, the network weights $\left \{ w^{l} \right \}_{l=0}^{L}$ and parameters $\left \{\lambda^{l}\right \}_{l=0}^{L}$ of the ANN are transformed into the weights and spiking fire thresholds $\left \{ \theta ^{l} \right \}_{l=0}^{L}$ in the SNN~\citep{deng2021optimal, deng2020rethinking}.

\subsubsection{Bio-information-fused training}
Although the A2S conversion demonstrates superior performance in previous work, there are still some inherent problems that are hard to avoid. The biology parameters, spiking firing threshold and membrane potential decay factor in LIF neuron, significantly influence the spatio-temporal information processing capability of SNN models~\citep{9760720, sun2023synapse, fang2021incorporating}. In A2S conversion method, these parameters are usually determined empirically or by threshold balancing strategies, limiting the performance of converted models.

To address these challenges, we propose a hybrid bio-information-fused training strategy~(BISNN). This approach incorporates two key operations: (1) Cross-model initialization: The SNN models, which share an identical architecture, are initialized with the pre-trained weights $\left \{ w^{l} \right \}_{l=0}^{L}$ of the ANN models. This facilitates efficient knowledge transfer between models; and (2) Parameter-free optimization: A supervised retraining procedure utilizing the STBP~\citep{wu2018spatio} method is employed to optimize the SNN models, thereby circumventing the need for complex biological parameter filtering processes. According to the chain rule, the mathematical formulation of the loss function's derivatives with respect to the learnable parameters $\left \{ w^{l} \right \}_{l=0}^{L}$ can be expressed as follows:
\begin{equation}
\frac{\partial \mathcal{L}_{\mathrm{flow}}}{\partial v^{l}(t)}=
\frac{\partial \mathcal{L}_{\mathrm{flow}}}{\partial s^{l}(t)}\frac{\partial s^{l}(t)}{\partial v^{l}(t)} 
+   \frac{\partial \mathcal{L}_{\mathrm{flow}}}{\partial s^{l}(t+1)}\frac{\partial s^{l}(t+1)}{\partial v^{l}(t)},
\label{eq15}
\end{equation}
\begin{equation}
\frac{\partial \mathcal{L}_{\mathrm{flow}}}{\partial w^{l}}=\sum_{t=1}^{T} 
\frac{\partial \mathcal{L}_{\mathrm{flow}}}{\partial v^{l}(t)}
\frac{\partial v^{l}(t)}{\partial w^{l}}.
\label{eq16}
\end{equation}
In this study, the approximate gradient function employed in the backpropagation process is formulated as follows~\citep{fang2021incorporating}:
\begin{equation}
H(\cdot )=\frac{\text{arctan}[\pi (\cdot )]}{\pi} + \frac{1}{2},
\label{eq17}
\end{equation}
where $\text{arctan}$ represents the inverse tangent function.

\begin{table}[!tb]\footnotesize
\caption{Comparison of event-based datasets for optical flow estimation. GT denotes the ground truth datasets. }
\label{table1}
\centering
\setlength{\tabcolsep}{1.8mm}{
	\begin{tabular}{l|ccccc}
		\hline
		Datasets                                   & Year &   Resolution   &   Train    &    Test    &     GT     \\ \hline
		UZH-FPV~\citep{zhu2018multivehicle}         & 2018 & 346$\times$260 & \checkmark &  $\times$  &  $\times$  \\
		ECD~\citep{mueggler2017event}               & 2017 & 240$\times$180 &  $\times$  & \checkmark &  $\times$  \\
		MVSEC~\citep{zhu2018multivehicle,zhu2018ev} & 2019 & 346$\times$260 &  $\times$  & \checkmark & \checkmark \\
		HQF~\citep{stoffregen2020reducing}          & 2020 & 240$\times$180 &  $\times$  & \checkmark &  $\times$  \\ \hline
\end{tabular}}
\end{table}

\begin{table*}[!htbp]\scriptsize
\caption{Comparison of state-of-the-art models across different datasets. The ST-FlowNet model trained using the ANN, A2S, and BISNN methods are denoted as ST-FlowNet\(_{1}\), ST-FlowNet\(_{2}\), and ST-FlowNet\(_{3}\), respectively. For the MVSEC dataset, the AEE results for each scenario are presented. For the ECD and HQF datasets, the average FWL and RSAT results across all scenarios are reported. The optimal and suboptimal results are highlighted in \textbf{bold}. The symbol $\downarrow$ ($\uparrow$) indicates that a smaller~(larger) value is preferred.}
\label{table2}
\centering
\setlength{\tabcolsep}{1.5mm}{
	\begin{tabular}{cl|cccccccc|cc|cc}
		\hline
             \multicolumn{2}{c|}{\multirow{2}{*}{Model}}                &          \multicolumn{2}{c}{OD1}          &          \multicolumn{2}{c}{IF1}          &          \multicolumn{2}{c}{IF2}          &         \multicolumn{2}{c|}{IF3}          &     \multicolumn{2}{c|}{ECD}     &     \multicolumn{2}{c}{HQF}      \\
                        \multicolumn{2}{c|}{}                           & AEE$_{1}\downarrow$ & AEE$_{4}\downarrow$ & AEE$_{1}\downarrow$ & AEE$_{4}\downarrow$ & AEE$_{1}\downarrow$ & AEE$_{4}\downarrow$ & AEE$_{1}\downarrow$ & AEE$_{4}\downarrow$ & FWL$\uparrow$ & RSAT$\downarrow$ & FWL$\uparrow$ & RSAT$\downarrow$ \\ \hline
		\multirow{7}{*}{ANN} & EV-FlowNet~\citep{hagenaars2021self}              &    \textbf{0.32}    &    \textbf{1.30}    &        0.58         &        2.18         &        1.02         &        3.85         &        0.87         &        3.18         &     1.31      &  \textbf{0.94}   &     1.37      &  \textbf{0.92}   \\
                    & RNN-EV-FlowNet~\citep{XU2025107447}               &         --          &        1.69         &         --          &    \textbf{2.02}    &         --          &    \textbf{3.84}    &         --          &    \textbf{2.97}    & \textbf{1.36} &       0.95       & \textbf{1.45} &       0.93       \\
                    & GRU-EV-FlowNet~\citep{hagenaars2021self}          &        0.47         &        1.69         &        0.60         &        2.16         &        1.17         &        3.90         &        0.93         &        3.00         &      --       &        --        &      --       &        --        \\
                    & GRU-FireNet~\citep{scheerlinck2020fast}           &        0.55         &        2.04         &        0.89         &        3.35         &        1.62         &        5.71         &        1.35         &        4.68         &      --       &        --        &      --       &        --        \\
                    & ET-FlowNet~\citep{tian2022event}                  &    \textbf{0.39}    &        1.47         &    \textbf{0.57}    &        2.08         &        1.20         &        3.99         &        0.95         &        3.13         &      --       &        --        &      --       &        --        \\
                    & STT-FlowNet~\citep{10.1007/978-3-031-78354-8_30}  &        0.66         &         --          &        0.57         &         --          &    \textbf{0.88}    &         --          &    \textbf{0.73}    &         --          &      --       &        --        &      --       &        --        \\
                    & ST-FlowNet$_1$~(ours)                             &        0.40         &    \textbf{1.24}    &    \textbf{0.48}    &    \textbf{1.86}    &    \textbf{0.89}    &    \textbf{2.98}    &    \textbf{0.70}    &    \textbf{2.34}    & \textbf{1.37} &  \textbf{0.92}   & \textbf{1.48} &  \textbf{0.90}   \\ \hline
		\multirow{9}{*}{SNN} & LIF-EV-FlowNet~\citep{hagenaars2021self}          &        0.53         &        2.02         &        0.71         &        2.63         &        1.44         &        4.93         &        1.16         &        3.88         &     1.21      &       0.95       &     1.24      &       0.94       \\
                    & XLIF-EV-FlowNet~\citep{hagenaars2021self}         &        0.45         &        1.67         &        0.73         &        2.72         &        1.45         &        4.93         &        1.17         &        3.91         &     1.23      &       0.95       &     1.25      &       0.93       \\
                    & LIF-FireNet~\citep{hagenaars2021self}             &        0.57         &        2.12         &        0.98         &        3.72         &        1.77         &        6.27         &        1.50         &        5.23         &     1.28      &       0.99       &     1.34      &       1.00       \\
                    & XLIF-FireNet~\citep{hagenaars2021self}            &        0.54         &        2.07         &        0.98         &        3.73         &        1.82         &        6.51         &        1.54         &        5.43         &     1.29      &       0.99       &     1.39      &       0.99       \\
                    & Adaptive-SpikeNet~\citep{10160551}                &        0.44         &         --          &        0.79         &         --          &        1.37         &         --          &        1.11         &         --          &      --       &        --        &      --       &        --        \\
                    & SDformerFlow~\citep{10.1007/978-3-031-78354-8_30} &        0.69         &         --          &        0.61         &         --          &        0.83         &         --          &    \textbf{0.76}    &         --          &      --       &        --        &      --       &        --        \\
                    & FSFN~\citep{apolinario2025stllr}                  &        0.50         &         --          &        0.76         &         --          &        1.19         &         --          &        1.00         &         --          &      --       &        --        &      --       &        --        \\
                    & ST-FlowNet$_2$~(ours)                             &    \textbf{0.37}    &    \textbf{1.24}    &    \textbf{0.50}    &    \textbf{1.86}    &    \textbf{0.84}    &    \textbf{2.78}    &    \textbf{0.70}    &    \textbf{2.34}    & \textbf{1.37} &  \textbf{0.91}   & \textbf{1.47} &  \textbf{0.90}   \\
                    & ST-FlowNet$_3$~(ours)                             &    \textbf{0.39}    &    \textbf{1.47}    &    \textbf{0.51}    &    \textbf{1.92}    &    \textbf{0.99}    &    \textbf{3.33}    &        0.77         &    \textbf{2.56}    & \textbf{1.34} &  \textbf{0.93}   & \textbf{1.45} &  \textbf{0.91}   \\ \hline
	\end{tabular}}
\end{table*}

\begin{table*}[!htbp]
\scriptsize
\centering
\caption{Comparison of models trained or fine-tuned on the MVSEC dataset.}
\label{table_add1}
\setlength{\tabcolsep}{9.4mm}{
\begin{tabular}{l|cccc}
	\hline
	Model                                      &       OD1        &       IF1        &       IF2        &       IF3        \\ \hline
	RNN-FireNet-S-FT~\citep{xu2025event}       & 1.97~(AEE$_{4}$) & 3.24~(AEE$_{4}$) & 5.48~(AEE$_{4}$) & 4.45~(AEE$_{4}$) \\
	STE-FlowNet~\citep{ding2022spatio}         & 0.42~(AEE$_{1}$) & 0.57~(AEE$_{1}$) & 0.79~(AEE$_{1}$) & 0.72~(AEE$_{1}$) \\
	U-Net-like SNN~\citep{cuadrado2023optical} &        --        & 0.58~(AEE$_{1}$) & 0.72~(AEE$_{1}$) & 0.67~(AEE$_{1}$) \\\hline
	\multirow{2}{*}{ST-FlowNet$_2$~(ours)}                      &  0.37~(AEE$_{1}$)   &  0.50~(AEE$_{1}$)   &  0.84~(AEE$_{1}$)   &  0.70~(AEE$_{1}$)   \\
	&  1.24~(AEE$_{4}$)   &  1.86~(AEE$_{4}$)   &  2.78~(AEE$_{4}$)   &  2.34~(AEE$_{4}$)   \\ \hline
\end{tabular}}
\end{table*}
\begin{figure*}[!t]
\centering
\includegraphics[width=6.8in]{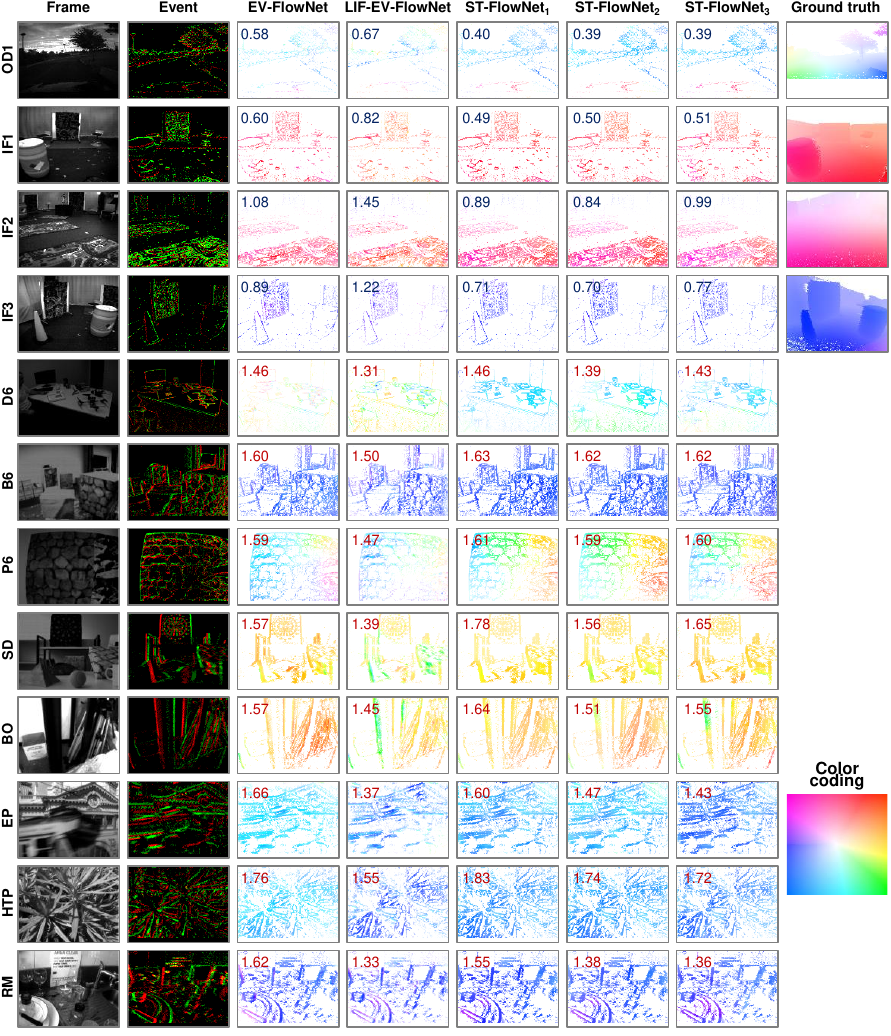}
\caption{Visual comparison of ST-FlowNet with other models. Original frame-based images, ground truth of the MVSEC dataset, and the color coding of the optical flow are provided for reference. The AEE$_{1}$ (black) or FWL (red) results of each predicted optical flow are provided at the upper-left.}
\label{Figure3}
\end{figure*}

%%%%%%%%%%%%%%%%%%%%%%%%%%%%%%%%%%%%%%%%% Experiments %%%%%%%%%%%%%%%%%%%%%%%%%%%%%%%%%%%%%%%%%
\section{Experiments}
In this section, we first provide a comprehensive overview of our experimental settings, encompassing datasets, evaluation metrics, and implementation details. Next, we present an in-depth performance comparison between ST-FlowNet and other state-of-the-art models on diverse benchmark datasets. For the purpose of visualization, we display representative examples for qualitative illustration. Finally, we conduct a series of ablation studies to demonstrate the significance of the proposed components. 

\subsection{Experimental Settings}
\subsubsection{Optical Flow Datasets}
We train the models using the UZH-FPV drone racing dataset~\citep{zhu2018multivehicle}, which is distinguished by a diverse distribution of optical flow vectors. We evaluate the models’ performance using the Event-Camera Dataset (ECD)~\citep{mueggler2017event}, Multi-Vehicle Stereo Event Camera (MVSEC)~\citep{zhu2018multivehicle,zhu2018ev}, and High-Quality Frames (HQF)~\citep{stoffregen2020reducing} dataset, all captured in real-world scenarios using various DAVIS neuromorphic cameras. Tab.~\ref{table1} shows a detailed overview of the dataset.

\begin{table*}[tb]\footnotesize
\caption{Performance analysis of the key modules in the ST-FlowNet architecture on the MVSEC dataset. The number of \textbf{Decoder} modules is denoted as '\#Ds', while the \textbf{ConvGRU1} and \textbf{ConvGRU2} modules are referred to as 'Aug' and 'Align', respectively. The invalid results is highlighted in $\underline{underline}$.}
\label{table3}
\centering
\setlength{\tabcolsep}{8.0mm}{
	\begin{tabular}{cccccccc}
		\hline
		\multirow{2}{*}{Method} & \multirow{2}{*}{\#Ds} & \multirow{2}{*}{Aug} & \multirow{2}{*}{Align} &          OD1           &          IF1           &          IF2           &          IF3           \\
		&                       &                      &                        & AEE$_{1}$$\downarrow $ & AEE$_{1}$$\downarrow $ & AEE$_{1}$$\downarrow $ & AEE$_{1}$$\downarrow $ \\ \cline{1-8}
		\multirow{4}{*}{ANN}   &           4           &       $\times$       &        $\times$        &          0.58          &          0.60          &          1.07          &          0.99          \\
		&           1           &       $\times$       &        $\times$        &          0.45          &          0.58          &          1.02          &          0.84          \\
		&           1           &     $\checkmark$     &        $\times$        &          0.44          &          0.53          &          0.99          &          0.81          \\
		&           1           &     $\checkmark$     &      $\checkmark$      &     \textbf{0.40}      &     \textbf{0.48}      &     \textbf{0.89}      &     \textbf{0.70}      \\ \cline{1-8}
		\multirow{4}{*}{A2S}  &           4           &       $\times$       &        $\times$        &          0.56          &          0.63          &          1.12          &          0.94          \\
		&           1           &       $\times$       &        $\times$        &          0.48          &          0.62          &          1.03          &          0.81          \\
		&           1           &     $\checkmark$     &        $\times$        &    \underline{0.53}    &          0.54          &          0.98          &          0.80          \\
		&           1           &     $\checkmark$     &      $\checkmark$      &     \textbf{0.37}      &     \textbf{0.50}      &     \textbf{0.84}      &     \textbf{0.70}      \\ \cline{1-8}
		\multirow{4}{*}{BISNN}  &           4           &       $\times$       &        $\times$        &          0.67          &          0.82          &          1.45          &          1.22          \\
		&           1           &       $\times$       &        $\times$        &          0.47          &          0.64          &          1.13          &          0.91          \\
		&           1           &     $\checkmark$     &        $\times$        &    \underline{0.49}    &          0.58          &          1.05          &          0.83          \\
		&           1           &     $\checkmark$     &      $\checkmark$      &     \textbf{0.39}      &     \textbf{0.51}      &     \textbf{0.99}      &     \textbf{0.77}      \\ \hline
\end{tabular}}
\end{table*}

\begin{table*}[tb]\footnotesize
\caption{Performance analysis of the key modules in the ST-FlowNet architecture on the ECD and HQF datasets. The number of \textbf{Decoder} modules is denoted as 'Ds', while the \textbf{ConvGRU1} and \textbf{ConvGRU2} modules are referred to as 'Aug' and 'Align', respectively.}
\label{table4}
\centering
\setlength{\tabcolsep}{3.5mm}{
	\begin{tabular}{ccccccccccccc}
		\hline
		\multirow{14}{*}{ECD} & \multirow{2}{*}{Method} & \multirow{2}{*}{\#Ds} & \multirow{2}{*}{Aug} & \multirow{2}{*}{Align} &        \multicolumn{2}{c}{D6}        &        \multicolumn{2}{c}{B6}        &        \multicolumn{2}{c}{P6}        &        \multicolumn{2}{c}{SD}        \\
		&                         &                       &                      &                        &  FWL$\uparrow $  & RSAT$\downarrow $ &  FWL$\uparrow $  & RSAT$\downarrow $ &  FWL$\uparrow $  & RSAT$\downarrow $ &  FWL$\uparrow $  & RSAT$\downarrow $ \\ \cline{2-13}
		&  \multirow{4}{*}{ANN}   &           4           &       $\times$       &        $\times$        &       1.29       &       0.90        &       1.51       &       0.92        &       1.45       &       0.92        &       1.40       &       0.92        \\
		&                         &           1           &       $\times$       &        $\times$        & \underline{1.45} &   \textbf{0.87}   &       1.62       &       0.92        &  \textbf{1.61}   &       0.92        &       1.61       &       0.90        \\
		&                         &           1           &     $\checkmark$     &        $\times$        &       1.42       & \underline{0.88}  &       1.62       &   \textbf{0.91}   & \underline{1.59} &   \textbf{0.91}   &       1.73       &   \textbf{0.89}   \\
		&                         &           1           &     $\checkmark$     &      $\checkmark$      &  \textbf{1.46}   &   \textbf{0.87}   &  \textbf{1.63}   & \underline{0.92}  &  \textbf{1.61}   &   \textbf{0.91}   &  \textbf{1.77}   &   \textbf{0.89}   \\ \cline{2-13}
		&  \multirow{4}{*}{A2S}   &           4           &       $\times$       &        $\times$        &       1.24       &       0.93        &       1.44       &       0.94        &       1.39       &       0.94        &       1.30       &       0.93        \\
		&                         &           1           &       $\times$       &        $\times$        &       1.40       &   \textbf{0.88}   &       1.60       &   \textbf{0.92}   &       1.57       &       0.92        &       1.51       &       0.91        \\
		&                         &           1           &     $\checkmark$     &        $\times$        &       1.40       & \underline{0.89}  &       1.61       &   \textbf{0.92}   &       1.59       &   \textbf{0.91}   &       1.56       &       0.91        \\
		&                         &           1           &     $\checkmark$     &      $\checkmark$      &  \textbf{1.44}   &   \textbf{0.88}   &  \textbf{1.63}   &   \textbf{0.92}   &  \textbf{1.62}   &   \textbf{0.91}   &  \textbf{1.62}   &   \textbf{0.90}   \\ \cline{2-13}
		& \multirow{4}{*}{BISNN}  &           4           &       $\times$       &        $\times$        &       1.31       &       0.90        &       1.50       &       0.93        &       1.47       &       0.93        &       1.39       &   \textbf{0.89}   \\
		&                         &           1           &       $\times$       &        $\times$        &       1.44       &   \textbf{0.88}   &       1.59       &       0.93        &       1.57       &       0.93        &       1.74       & \underline{0.91}  \\
		&                         &           1           &     $\checkmark$     &        $\times$        &       1.44       &   \textbf{0.88}   &       1.61       &   \textbf{0.91}   &       1.59       &   \textbf{0.91}   &  \textbf{1.83}   &   \textbf{0.89}   \\
		&                         &           1           &     $\checkmark$     &      $\checkmark$      &  \textbf{1.45}   &   \textbf{0.88}   &  \textbf{1.62}   & \underline{0.92}  &  \textbf{1.61}   &   \textbf{0.91}   & \underline{1.66} &   \textbf{0.89}   \\ \hline
		\multirow{14}{*}{HQF} & \multirow{2}{*}{Method} & \multirow{2}{*}{\#Ds} & \multirow{2}{*}{Aug} & \multirow{2}{*}{Align} &        \multicolumn{2}{c}{BO}        &        \multicolumn{2}{c}{EP}        &       \multicolumn{2}{c}{HTP}        &        \multicolumn{2}{c}{RM}        \\
		&                         &                       &                      &                        &  FWL$\uparrow $  & RSAT$\downarrow $ &  FWL$\uparrow $  & RSAT$\downarrow $ &  FWL$\uparrow $  & RSAT$\downarrow $ &  FWL$\uparrow $  & RSAT$\downarrow $ \\ \cline{2-13}
		&  \multirow{4}{*}{ANN}   &           4           &       $\times$       &        $\times$        &       1.37       &       0.91        &       1.29       &       0.93        &       1.49       &       0.90        &       1.24       &       0.94        \\
		&                         &           1           &       $\times$       &        $\times$        &       1.57       &   \textbf{0.90}   &       1.57       &   \textbf{0.90}   &       1.79       &       0.89        &       1.46       &       0.91        \\
		&                         &           1           &     $\checkmark$     &        $\times$        &  \textbf{1.65}   &   \textbf{0.90}   &       1.58       &       0.91        &       1.81       &   \textbf{0.88}   &       1.52       &       0.91        \\
		&                         &           1           &     $\checkmark$     &      $\checkmark$      &  \textbf{1.65}   &   \textbf{0.90}   &  \textbf{1.60}   &       0.91        &  \textbf{1.82}   &   \textbf{0.88}   &  \textbf{1.54}   &   \textbf{0.90}   \\ \cline{2-13}
		&  \multirow{4}{*}{A2S}   &           4           &       $\times$       &        $\times$        &       1.30       &       0.93        &       1.24       &       0.95        &       1.37       &       0.92        &       1.20       &       0.95        \\
		&                         &           1           &       $\times$       &        $\times$        &       1.46       &       0.91        &       1.46       &   \textbf{0.91}   &       1.64       &   \textbf{0.89}   &       1.35       &       0.92        \\
		&                         &           1           &     $\checkmark$     &        $\times$        &       1.58       &       0.91        &       1.58       &   \textbf{0.91}   &       1.79       &   \textbf{0.89}   &       1.48       &       0.91        \\
		&                         &           1           &     $\checkmark$     &      $\checkmark$      &  \textbf{1.60}   &   \textbf{0.90}   &  \textbf{1.59}   &   \textbf{0.91}   &  \textbf{1.79}   & \underline{0.90}  &  \textbf{1.49}   &   \textbf{0.90}   \\ \cline{2-13}
		& \multirow{4}{*}{BISNN}  &           4           &       $\times$       &        $\times$        &       1.45       &   \textbf{0.89}   &       1.37       &   \textbf{0.91}   &       1.55       &   \textbf{0.88}   &       1.33       &       0.92        \\
		&                         &           1           &       $\times$       &        $\times$        &       1.56       & \underline{0.91}  &       1.57       & \underline{0.92}  &       1.67       & \underline{0.91}  &       1.52       &       0.92        \\
		&                         &           1           &     $\checkmark$     &        $\times$        &       1.56       &       0.91        &  \textbf{1.59}   &   \textbf{0.91}   &       1.71       &       0.89        &  \textbf{1.53}   &   \textbf{0.91}   \\
		&                         &           1           &     $\checkmark$     &      $\checkmark$      &  \textbf{1.57}   &       0.90        & \underline{1.56} &   \textbf{0.91}   &  \textbf{1.77}   &   \textbf{0.88}   & \underline{1.49} &   \textbf{0.91}   \\ \hline
\end{tabular}}
\end{table*}

\subsubsection{Evaluation Metrics}
For the MVSEC~\citep{zhu2018multivehicle} optical flow dataset, the ground truth is generated at each APS frame timestamp and scaled to represent the displacement for the duration of one~(dt=1) and four~(dt=4) APS frames~\citep{zhu2018ev}. Consequently, optical flow is also computed at each APS frame timestamp, using all events within the time window as input for dt=1, or $25\%$ of the window events at a time for dt=4. Both predicted optical flows are evaluated using the average endpoint error~(AEE$_{1}$ for dt=1 and AEE$_{4}$ for dt=4).

The ECD and HQF datasets do not include ground truth, and thus we employ two image compensation quality metrics to estimate predicted flows. Specifically, the flow warp loss~(FWL) assesses the sharpness of the image of warped events compared with the original event partition, and the variance of the contrast of the event images is reported~\citep{Stoffregen20eccv}. Additionally, we also report the ratio of the squared average timestamps~(RSAT), indicating the contrast of $\mathcal{L}_{\mathrm{contrast}}$ between predicted optical flow and baseline null vectors~\citep{hagenaars2021self}.

\subsubsection{Implementation Details}
We implement ST-FlowNet using the PyTorch framework and execute on an NVIDIA A100 GPU. The training process consists of 100 epochs for the ANN model, followed by 10 epochs of retraining in the BISNN method, with a batch size of 8. We use the adaptive moment estimation optimizer~\citep{kingma2014adam} with an initial learning rate of 0.0002, subject to exponential decay. We empirically set the scaling weight for $\mathcal{L}_{\mathrm{smooth}}$ to $\lambda = 0.001$, as used in previous work~\citep{zhu2018ev}. 

In the SNN models converted from ANN models, the firing thresholds are determined using the threshold balance strategy. To simplify the parameter selection process, we focus exclusively on the membrane potential decay factors in the \textbf{Generator}~($\tau_{0}$) and \textbf{ConvGRU2}~($\tau_{1}$) modules, while setting those in other modules to 0. For consistency and a fair comparison, during the retraining procedure of the BISNN method, both the firing threshold and membrane potential decay factors are initialized with the same values as those used in the A2S method.

\subsection{Comparison with State-of-the-art Methods}
To validate the effectiveness of our proposed ST-FlowNet, we conduct a comprehensive comparative analysis, assessing its performance compared with other state-of-the-art models from both quantitative and qualitative perspectives~\citep{scheerlinck2020fast,tian2022event}. The ST-FlowNet model trained using the ANN, A2S, and BISNN methods are denoted as ST-FlowNet\(_{1}\), ST-FlowNet\(_{2}\), and ST-FlowNet\(_{3}\), respectively.

\subsubsection{Quantitative Evaluation}
We conduct a comparative analysis between ST-FlowNet and other state-of-the-art models on the MVSEC dataset across four representative scenarios: outdoor\_day1~(denoted as OD1), indoor\_flying1~(IF1), indoor\_flying2~(IF2) and indoor\_flying3~(IF3), considering both the dt=1 and dt=4 conditions. As shown in Tab.~\ref{table2}, ST-FlowNet demonstrates superior AEE performance compared with other ANN and SNN models across most of scenarios. Notably, the optical flow estimation performance can be improved slightly by leveraging the converted SNN model, as particularly evident in the OD1 and IF2 scenarios. We also provide a comparison between our ST-FlowNet$_{2}$ SNN model and other models that are directly trained or fine-tuned on the MVSEC dataset. As shown in Tab.~\ref{table_add1}, ST-FlowNet$_{2}$ still demonstrates competitive performance compared to other models. We believe that training on a larger dataset may further enhance the optical flow estimation performance of our model. For the sake of comparison, on the ECD and HQF dataset, we present a comparative analysis of state-of-the-art models without distinguishing scenarios. As shown in Tab.~\ref{table2}, ST-FlowNet attains the optimal performance, excelling with respect to both the FWL and RSAT metrics.

\subsubsection{Qualitative Evaluation}
In Fig.~\ref{Figure3}, we present the visualization results of the predicted optical flows generated by various models. We use EV-FlowNet~(ANN model) and LIF-EV-FlowNet~(SNN model), which exhibit outstanding performance, for comparison with our ANN and SNN models, respectively. The AEE$_{1}$~(black) or FWL~(red) values are provided at the upper-left of each predicted optical flow image. Overall, ST-FlowNet demonstrates evident superiority in the visual quality of its optical flows compared with competing models. Specifically, in challenging scenarios such as boundary regions with motion blur or sparse features~(i.e. D6 and HTP), both EV-FlowNet and LIF-EV-FlowNet yield prediction failures or errors. ST-FlowNet excels in accurately capturing detailed scene information, thereby achieving reliable optical flow estimation. Additionally, qualitative comparisons highlight that the converted SNN model retains the original ANN model's proficiency in optical flow estimation.

\begin{figure*}[t]
\centering
\includegraphics[width=7.2in]{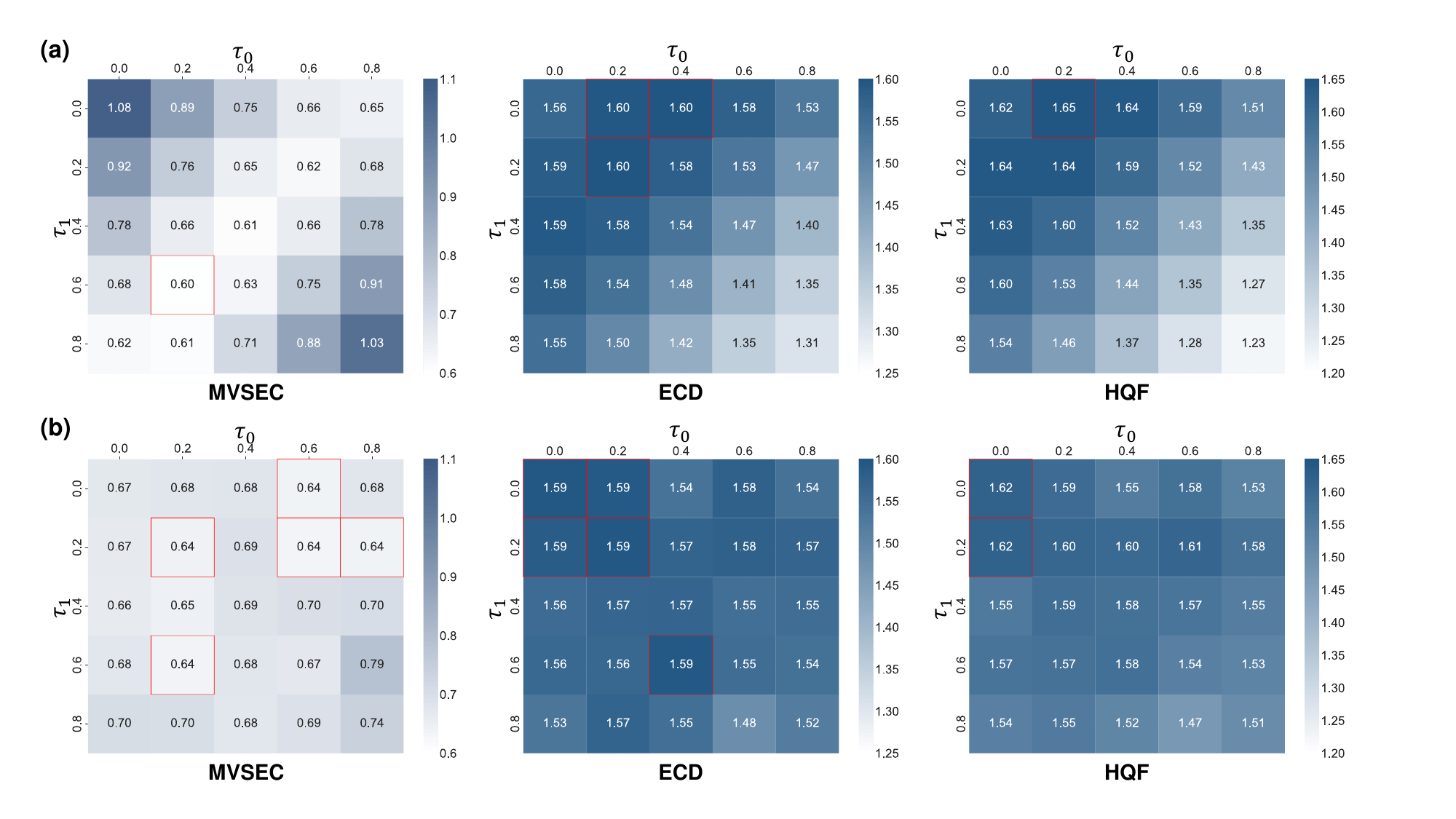}
\caption{The performance comparison of SNN models initialized with different combinations of membrane potential decay factors. (a) SNN models trained using the A2S method. (b) SNN models trained using the BISNN method. The optimal results are highlighted within red boxes.}
\label{Figure4}
\end{figure*}

\subsection{Ablation Analysis}
Our novel ST-FlowNet architecture comprises the following principal components: decoders with a non-pyramid architecture~(\textbf{Decoder}), spatio-temporal optical flow augmentation~ (\textbf{ConvGRU1}), and optical flow alignment~(\textbf{ConvGRU2}). To demonstrate the effectiveness of these components, we establish a baseline architecture by removing two ConvGRU layers and 
retaining only one decoder in ST-FlowNet. We conduct relative ablation analyses on the baseline architecture by progressively introducing the relevant components.

\subsubsection{Analysis of the Number of Decoders}
We compare the baseline models with varying numbers of decoders to examine the effectiveness of the decoders. Specifically, we increase the number of decoders from one in the baseline model to four in the comparison models. To conduct a thorough comparison, we select four representative scenarios from the ECD and HQF datasets, each denoted as follows: dynamic\_6dof (D6), boxes\_6dof (B6), poster\_6dof (P6), slider\_depth (SD), along with boxes~(BO), engineering\_posters (EP), high\_texture\_plants (HTP), and reflective\_materials (RM). As shown in Tabs.~\ref{table3} and~\ref{table4}, the performance decreases across all scenarios when more decoders are employed. This suggests that the use of more decoders is not necessarily associated with improved performance. This observation not only alleviates the pressure to increase the model size but also reinforces the advantage of the SNN model in terms of energy consumption.

\subsubsection{Analysis of the Spatio-temporal Augmentation}
Building on the baseline model with a single decoder, we integrate \textbf{ConvGRU1} as the spatio-temporal optical flow augmentation component. Unlike the complete ST-FlowNet, the \textbf{ConvGRU1} layer receives the upsampled basic predicted optical flow as state information. The detailed results are shown in Tabs.~\ref{table3} and~\ref{table4}. The AEE results for the MVSEC dataset demonstrate that the ConvGRU augmentation layer enhances optical flow estimation performance in most scenarios, except for OD1. The FWL and RSAT results computed for the ECD and HQF datasets also generally corroborate the improvement, which is more pronounced for the HQF dataset. Furthermore, the ConvGRU layers assist in mitigating the performance loss of converted SNN models. In certain scenarios, SNN models exhibit superior performance compared with ANN models, as observed for the IF2 and IF3 scenarios.

\subsubsection{Analysis of the Spatio-temporal Alignment}
Expanding on the previous model, we continue to introduce \textbf{ConvGRU2} as a spatio-temporal alignment module. This module is designed to project the basic and historical predicted optical flow into a standardized state space. By doing so, it not only enhances the precision of optical flow prediction but also establishes an effective reference state for input data. As illustrated in Tabs.~\ref{table3} and~\ref{table4}, the optical flow estimation performance further improves in most scenarios. Moreover, in comparison to the spatio-temporal augmentation module, the alignment module demonstrates superior effectiveness.

\begin{table}[tb]\footnotesize
\caption{Performance analysis of SNNs trained with different method.}
\label{table5}
\centering
\setlength{\tabcolsep}{0.7mm}{
	\begin{tabular}{cccccccccc}
		\hline
		\multirow{5}{*}{\rotatebox{90}{MVSEC}} & \multirow{2}{*}{Method} &          \multicolumn{2}{c}{OD1}           &          \multicolumn{2}{c}{IF1}           &          \multicolumn{2}{c}{IF2}           &          \multicolumn{2}{c}{IF3}           \\
		&                         & \multicolumn{2}{c}{AEE$_{1}$$\downarrow $} & \multicolumn{2}{c}{AEE$_{1}$$\downarrow $} & \multicolumn{2}{c}{AEE$_{1}$$\downarrow $} & \multicolumn{2}{c}{AEE$_{1}$$\downarrow $} \\ \cline{2-10}
		&          STBP           &          \multicolumn{2}{c}{0.78}          &          \multicolumn{2}{c}{1.00}          &          \multicolumn{2}{c}{1.78}          &          \multicolumn{2}{c}{1.57}          \\
		&          A2S           &          \multicolumn{2}{c}{0.37}          &          \multicolumn{2}{c}{0.50}          &          \multicolumn{2}{c}{0.84}          &          \multicolumn{2}{c}{0.70}          \\
		&          BISNN          &     \multicolumn{2}{c}{\textbf{0.39}}      &     \multicolumn{2}{c}{\textbf{0.51}}      &     \multicolumn{2}{c}{\textbf{0.99}}      &     \multicolumn{2}{c}{\textbf{0.77}}      \\ \hline
		\multirow{5}{*}{\rotatebox{90}{ECD}}  &                         &           \multicolumn{2}{c}{D6}           &           \multicolumn{2}{c}{B6}           &           \multicolumn{2}{c}{P6}           &           \multicolumn{2}{c}{SD}           \\
		&                         & FWL$\uparrow $ &     RSAT$\downarrow $     & FWL$\uparrow $ &     RSAT$\downarrow $     & FWL$\uparrow $ &     RSAT$\downarrow $     & FWL$\uparrow $ &     RSAT$\downarrow $     \\ \cline{2-10}
		&          STBP           &      1.18      &           0.95            &      1.31      &           0.96            &      1.30      &           0.96            &      1.46      &           0.92            \\
		&          A2S           &      1.44      &       \textbf{0.88}       & \textbf{1.63}  &       \textbf{0.92}       & \textbf{1.62}  &       \textbf{0.91}       &      1.62      &           0.90            \\
		&          BISNN          & \textbf{1.45}  &       \textbf{0.88}       &      1.62      &       \textbf{0.92}       &      1.61      &       \textbf{0.91}       & \textbf{1.66}  &       \textbf{0.89}       \\ \hline
		\multirow{5}{*}{\rotatebox{90}{HQF}}  &                         &           \multicolumn{2}{c}{BO}           &           \multicolumn{2}{c}{EP}           &          \multicolumn{2}{c}{HTP}           &           \multicolumn{2}{c}{RM}           \\
		&                         & FWL$\uparrow $ &     RSAT$\downarrow $     & FWL$\uparrow $ &     RSAT$\downarrow $     & FWL$\uparrow $ &     RSAT$\downarrow $     & FWL$\uparrow $ &     RSAT$\downarrow $     \\ \cline{2-10}
		&          STBP           &      1.34      &           0.93            &      1.16      &           0.97            &      1.44      &           0.93            &      1.17      &           0.96            \\
		&          A2S           & \textbf{1.60}  &       \textbf{0.90}       & \textbf{1.59}  &       \textbf{0.91}       & \textbf{1.79}  &           0.90            & \textbf{1.49}  &       \textbf{0.90}       \\
		&          BISNN          &      1.57      &       \textbf{0.90}       &      1.56      &       \textbf{0.91}       &      1.77      &       \textbf{0.88}       & \textbf{1.49}  &           0.91            \\ \cline{1-10}
\end{tabular}}
\end{table}

\subsection{Analysis of Training Methods}
To thoroughly assess the efficacy of various training approaches, we also conduct experiments involving the direct training of SNNs using the STBP method. The initialization operations of biological parameters are consistent with those in the BISNN method. As presented in Tab.~\ref{table5}, the performance of models trained using the A2S and BISNN methods significantly outperforms those trained with the STBP method. These findings suggest that direct training of SNNs for optical flow estimation remains a challenging task, and further highlight the competitive performance improvements achieved by both the A2S and BISNN methods.

The performance of SNN models converted using the A2S method is influenced by the selection of biological parameters. In our experiments, the spike firing thresholds are determined using the threshold balance strategy, while the membrane potential decay factors are chosen empirically. To further verify the impact of parameter settings on model performance, we perform a search over the parameters $\tau_{0}$ and $\tau_{1}$ within the $[0,0.8]$ range. Experiments are conducted on models trained using both the A2S and BISNN methods under various parameter combinations, and the average results across all scenarios are presented in Fig.~\ref{Figure4}. The optimal results for each set of experiments are highlighted within red boxes.

As shown in Fig.~\ref{Figure4}(a), the performance of the models fluctuates significantly with changes in the membrane potential decay factors. On the MVSEC dataset, parameter combinations along the secondary diagonal tend to yield the best results. However, for the ECD and HQF datasets, smaller membrane potential decay factor combinations appear to be more effective. Overall, we are still unable to summarize a general principle for parameter selection from this sufficiently large search space. These findings underscore the challenges associated with selecting optimal biological parameters. Nevertheless, the performance of models trained using the BISNN method demonstrates notable robustness across varying parameter combinations. These results substantiate our claim that BISNN operates as a parameter-free training method, effectively alleviating the complexities of parameter tuning while consistently delivering robust optical flow estimation performance.

\begin{figure}[t]
\centering
\includegraphics[width=3.4in]{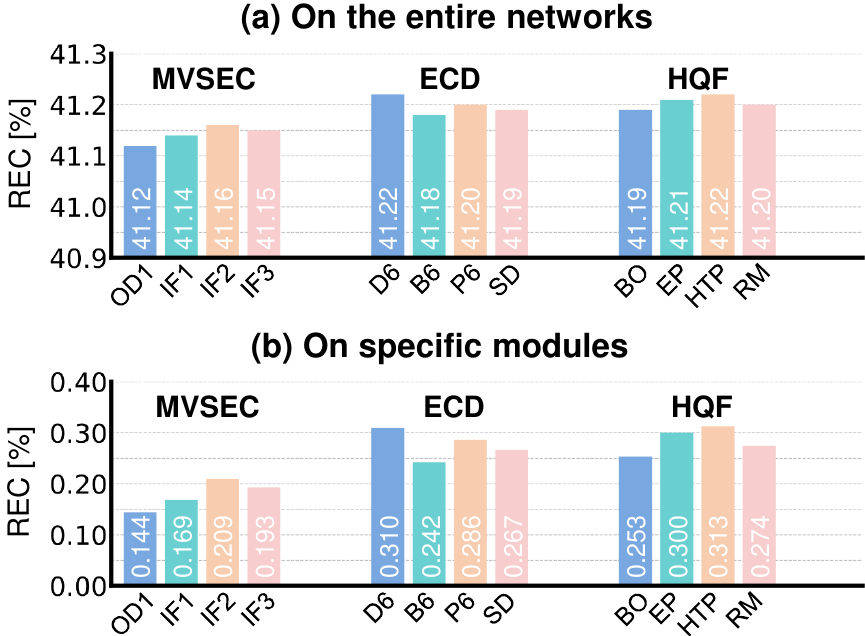}
\caption{The energy consumption of SNN ST-FlowNet models relative to ANN models.}
\label{Figure5}
\end{figure}

\subsection{Analysis of Energy Consumption}
Finally, we assess the energy consumption for different ST-FlowNet models. In contrast to the ANN ST-FlowNet model that mainly employs multiply-accumulate operations, the SNN ST-FlowNet model predominantly leverages sparse accumulate~(AC) operations as computational units, but excluding the \textbf{ConvGRU1/2}, \textbf{Encoder1} and \textbf{Decoder1} modules. The relative energy consumption~(REC), denoted as $\eta = \frac{\Phi_{\text{SNN}}}{\Phi_{\text{ANN}}}$, is employed to evaluate the energy-saving advantages of SNN ST-FlowNet models. Here, $\Phi$ represents the theoretical energy consumption as calculated in previous work~\citep{yao2023sparser}. 

We conduct experiments on ST-FlowNet models trained using the ANN and A2S methods, respectively, and the results are summarized in Fig.~\ref{Figure5}. As expected, the energy consumption reduction achieved by the SNN ST-FlowNet model is approximately $60\%$ across the entire networks~(Fig.~\ref{Figure5}~(a)). If we focus only on specific modules that exclusively use AC operations in the SNN ST-FlowNet model, a remarkable reduction~(more than 300-fold) in energy consumption can be observed~(Fig.~\ref{Figure5}~(b)). These observations clearly demonstrate the advantage of energy consumption in SNN ST-FlowNet models.

\begin{figure}[t]
\centering
\includegraphics[width=3.5in]{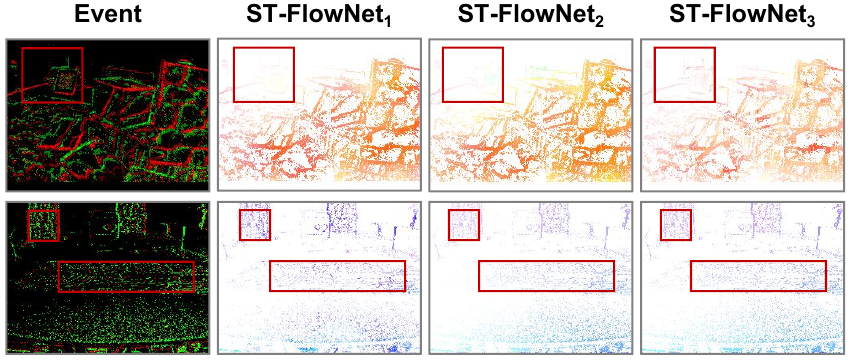}
	\caption{Representative examples of failure cases from our ST-FlowNet models. The regions with failed optical flow estimations are highlighted with red boxes.}
	\label{Figure6}
\end{figure}

\subsection{Analysis of Failure Cases}
Although our proposed method demonstrates promising performance in optical flow estimation, several challenges remain unresolved. Two representative failure cases are illustrated in Fig.~\ref{Figure6}. In the first row, the ST-FlowNet models fail to accurately capture a small object, as highlighted by the red box. This observation is consistent with the visualization results in Fig.~\ref{Figure3}, which indicate a relatively degraded performance in scenarios containing numerous small-scale objects (i.e. D6 and RM). In addition, distinguishing between background noise and high-density event streams presents another significant challenge, as shown in the second row of Fig.~\ref{Figure6}. This issue, common in event-based vision tasks, may potentially be addressed through the development of more suitable event representation strategies.

%%%%%%%%%%%%%%%%%%%%%%%%%%%%%%%%%%%%%%%%% Conclusion %%%%%%%%%%%%%%%%%%%%%%%%%%%%%%%%%%%%%%%%%
\section{Discussion and Conclusion}
In this study, we present a novel neural network architecture, termed ST-FlowNet, which incorporates enhanced ConvGRU layers, designed to align and augment spatio-temporal information, thereby improving optical flow estimation based on event-driven inputs. To address the challenges in training SNN models for optical flow estimation, we propose two methods: (1) the A2S method, which generates an SNN model from a pre-trained ANN model, and (2) a novel BISNN strategy, aimed at mitigating the complexities associated with the selection of biological parameters.

Overall, our work demonstrates a notable level of superiority. (1) Our experimental results across a variety of representative scenarios validate the effectiveness of the proposed ST-FlowNet, which outperforms current state-of-the-art optical flow estimation models. In particular, for the challenging boundary regions characterized by motion blur or sparse features~(commonly introduced during event representation), ST-FlowNet exhibits a robust capability for capturing scene information. (2) Ablation studies further highlight the critical roles of the integrated \textbf{ConvGRU} modules in spatio-temporal augmentation and alignment, establishing a promising tool for future model design. Meanwhile, increasing the number of decoder layers is shown to be non-essential in semi-pyramidal architectural models, thereby contributing to a lightweight and energy-efficient design. (3) A comparative analysis of three training paradigms demonstrates that indirect A2S conversion and hybrid BISNN methods can generate superior SNN models compared to direct STBP training. Notably, the BISNN method exhibits greater robustness to variations in initialized biological parameters, thereby alleviating the complexity of biological parameter selection.

The limitations and shortcomings of the ST-FlowNet model are discussed below. First, we observe performance degradation in the converted SNN models under specific conditions (e.g., IF1 in Tab.~\ref{table2} and HTP in Tab.~\ref{table4}), which may be attributed to theoretical conversion errors inherent in the A2S process~\citep{deng2021optimal}. This issue could be addressed through the development of more efficient conversion methods. Second, the spatio-temporal augmentation module exhibits a less robust ability to improve performance, potentially due to modality bias between the input event data and the optical flow state. A well-designed \textbf{ConvGRU1} module may further improve the performance of the ST-FlowNet model by reducing modality bias, deserving to be examined in our further studies. Third, analysis of the failure cases reveals that the ST-FlowNet models exhibit limited sensitivity to small objects, background noise, and high-intensity event streams. Addressing these challenges represents a potential avenue for further enhancing model performance.

%%%%%%%%%%%%%%%%%%%%%%%%%%%%%%%%%%%%%%%%% Acknowledgments %%%%%%%%%%%%%%%%%%%%%%%%%%%%%%%%%%%%%%%%%
\section*{Acknowledgments}
This work was supported in part by the National Key Research and Development Program of China under grant 2023YFF1204200, in part by the STI 2030–Major Projects under grant 2022ZD0208500, in part by the Sichuan Science and Technology Program under grant 2024NSFJQ0004 and grant DQ202410, and in part by the Natural Science Foundation of Chongqing, China under grant CSTB2024NSCQ-MSX0627, in part by the Science and Technology Research Program of Chongqing Education Commission of China under Grant KJZD-K202401603, and in part by the China Postdoctoral Science Foundation under grant 2024M763876.

%%%%%%%%%%%%%%%%%%%%%%%%%%%%%%%%%%%%%%%%%%%%%%%%%%%% References %%%%%%%%%%%%%%%%%%%%%%%%%%%%%%%%%%%%%%%%%%%%%%%%%%%%%%%%%%
\bibliographystyle{elsarticle-harv} 
\bibliography{ref}

\end{document}